\documentclass[11pt,letterpaper,final]{article}
\usepackage[latin1]{inputenc}
\usepackage[margin=1in,foot=0.5in,nohead]{geometry}
\usepackage{times}

\usepackage{amsmath}
\usepackage{amsfonts}
\usepackage{amssymb}
\usepackage{amsthm}
\usepackage{array}
\theoremstyle{plain}
\newtheorem{theorem}{Theorem}[section]

\newtheorem{remark}[theorem]{Remark}
\newtheorem{definition}[theorem]{Definition}

\newtheorem{example}[theorem]{Example}
\usepackage{graphicx}
\usepackage{multicol}
\usepackage{epsfig}
\usepackage{subfig}
\usepackage{algorithmic}
\usepackage{algorithm}
\usepackage{url}

\begin{document}

\title{Hamiltonian Streamline Guided Feature Extraction with Applications to Face Detection}
\author{Yingjie Miao \\ Department of Mathematics \\ State University of New York at Buffalo \\ ymiao@buffalo.edu \and Jason J. Corso \\ Department of Computer Science and Engineering \\ State University of New York at Buffalo\\jcorso@buffalo.edu}
\date{}

\maketitle

\begin{abstract}
We propose a new feature extraction method based on two dynamical systems induced by intensity landscape: the negative gradient system and the Hamiltonian system. We build features based on the Hamiltonian streamlines. These features contain nice global topological information about the intensity landscape, and can be used for object detection. We show that for training images of same size, our feature space is much smaller than that generated by Haar-like features. The training time is extremely short, and detection speed and accuracy is similar to Haar-like feature based classifiers.
\end{abstract}

\begin{section}{Introduction}
Feature extraction is a critical component in pattern recognition and image processing. In image pattern recognition, one usually designs a family of features, which provides a mapping from image space to feature space. If the features are well designed, one can expect that images from different classes of objects (say, human faces and cars) are well separated in the feature space. Abundant types of features exist, e.g. Haar-like features \cite{viola2002robust}, interest points operators \cite{lowe1999object}, template matching \cite{torralba2004sharing}, etc. However, many of these features are computationally expensive. For example, one popular method for object detection is design a large family of Haar-like features and use boosting method for feature selection. Consider an $n \times n$ training image, the number of all possible Haar-like features is of order $\Omega (n^4)$. In practice, one only affords to select a fraction of these features, so the resulting subset is still of order $\Omega (n^4)$ asymptotically. Since the complexity of a boosting method is proportional to the number of features, training process is usually slow, despite the fact that we can evaluate single feature value fast by using integral image.

In this paper, we propose a new feature extraction methodology. Our philosophy is that in many cases, we can do object detection by inspecting shape characteristics of the intensity landscape. For example, in a typical image landscape of human face, eyes and mouth can be identified as valleys, nose and cheeks as peaks. If we can find an efficient way encoding this landscape, we can, up to diffeomorphisms, recover the whole intensity landscape by decoding. Y.Shinagawa \cite{shinagawa2002surface} proposed a surface coding method based on Morse theory, which represents a closed 2-dimensional manifold by its Reeb graph. A major result in Morse theory states that the global shape of a closed manifold is completed determined (up to diffeomorphisms) by local critical points of a Morse function. This provides a powerful tool to infer global information from a few local information. However, Morse theory only deals with topological invariants, thus, much geometric information is lost in the resulting Reeb graph. (We refer readers to \cite{matsumoto2002introduction} for a complete introduction on Morse theory). In this paper, we propose a method that can capture both topological and geometric features of such landscapes. Our method is based on two dynamical systems induced by the intensity landscape: a negative gradient flow system and a Hamiltonian system. By applying results in dynamical systems,  we are able to capture the characteristics of the surface by a family of contours and the gradient information along the contours. It turns out that, empirically, there are about $O(n)$ such features for $n \times n$ training images. Our experiments on face detection shows that one can build an AdaBoost classifier \cite{viola2002robust} based on these features and the resulting classifier has high testing accuracy. Most importantly,  since the feature space is much smaller comparing to that of Haar-like features, training time is significantly shorter.

It is worthwhile to note that our feature extraction method, when applied to face detection, is far different from existing methods. Several popular existing methods are: 1) template matching \cite{yuille1991deformable,jesorsky2001robust}; 2) PCA based methods \cite{turk1991eigenfaces}; 3) Neural network based methods \cite{rowley2002neural}; 4) Haar-like features with AdaBoost algorithm \cite{viola2002robust}. See \cite{senior2002face} for a survey. The first type of method typically requires manually designed templates or large amount of manually labeled landmark points. The last three types of methods, although fully automated, are computationally intensive. Our algorithm has merits from both sides: it can intelligently capture a few face related feature candidates and do feature selection on a relatively small feature space, therefore achieve very fast automatic training.

The rest of this paper is organized as follows: in Section 2 we review some background of dynamical system theory. In Section 3 and 4, we describe algorithms for feature extraction. This is the main contribution of this paper. In Section 5 we show the general framework of our training process and some experimental results. In Section 6 we discuss limitations of our method and possible future work.

\end{section}

\begin{section}{Review of Dynamical Systems}
Any smooth function defined on the plane can be visualized as a landscape. This landscape naturally induces a gradient flow system, whose solution curves reflect geometric properties of the underlying landscape. An image can be viewed as a discretized landscape, which admits a discretized gradient flow. Our basic philosophy is to actively capture features in the image by studying geometric and topological properties of the flow. First, we provide a brief review of related topics in dynamical systems. For more information, we refer readers to \cite{hirsch2004differential} and \cite{guckenheimer2002nonlinear}.

\begin{subsection}{Gradient System and Hamiltonian System.}
A \textbf{planar dynamical system} is a system of differential equations:
\begin{equation} \label{eq:generalSystem}
\left\{
\begin{array}{rl}
\dot{x} & = f(x,y) \\
\dot{y} & = g(x,y) 
\end{array} \right.
\end{equation}
where $x\in \mathbb{R}$ , $y \in \mathbb{R}$, $f(x,y)$ and $g(x,y)$ are continuous real-valued functions. The dot means differentiate with respect to time $t$. A curve $\gamma(t) = ( X(t),Y(t)) $ is a \textbf{solution curve} of \eqref{eq:generalSystem}, if $\dot{X}(t) = f(X(t),Y(t))$ and $\dot{Y}(t)= g(X(t),Y(t))$ for any $t$ where $\gamma(t)$ is defined. Note \eqref{eq:generalSystem} defines a velocity field on plane, which generates a \textbf{flow} $\phi_t: \mathbb{R}^2 \rightarrow \mathbb{R}^2 $, where $\phi_t(x_0,y_0) $ can be obtained by starting at point $(x_0,y_0)$ and follow the velocity field for time $t$. If we replace $f(x,y)$ and $ g(x,y)$ by $-f(x,y)$ and $-g(x,y)$ respectively, we get a \textbf{backward flow} of \eqref{eq:generalSystem}.  \eqref{eq:generalSystem} is called  a \textbf{gradient system} if there is a function $ H(x,y)$ such that $f(x,y) = \frac{\partial H}{\partial x}$ and $g(x,y) = \frac{\partial H}{\partial y} ~$. Conversely, given $H(x,y)$, one can define a \textbf{negative gradient system}
\begin{equation} \label{eq:negGradient}
\left\{
\begin{array}{rl}
\dot{x} & = - \frac{\partial H}{\partial x} \\
\dot{y} & = -\frac{\partial H}{\partial y} 
\end{array} \right.
\end{equation}

Dynamical behavior of \eqref{eq:negGradient} is straightforward: at each point, the vector field points to the direction where $H(x,y)$ decays fastest. Therefore, local minima of $H$ become sinks in the system, whereas local maxima become sources. 

Given a gradient flow system, there exists a \textbf{Hamiltonian system} associated with it. For the same $H(x,y)$, one can define
\begin{equation} \label{eq:HamiFlow}
\left\{
\begin{array}{rl}
\dot{x} & = - \frac{\partial H}{\partial y} \\
\dot{y} & = \frac{\partial H}{\partial x} 
\end{array} \right.
\end{equation}
An important property is that solution curves of \eqref{eq:HamiFlow} are exactly level curves of $H$. To see this, let $ (x(t),y(t)) $ be any solution, we have 
\[ \frac{d}{dt}(H(x(t),y(t))) = \frac{\partial H}{\partial x} \frac{dx}{dt} + \frac{\partial H}{\partial y} \frac{dy}{dt} \equiv 0 \quad \quad  \forall t \]
It is also worthwhile to note that solution curves of the gradient system and those of the Hamiltonian system form an orthogonal family, because the two vector fields are perpendicular to each other at every nonsingular point. 

In Figure \ref{fig:landscape}, we show an average face image, its intensity landscape, and the associated negative gradient flow and the Hamiltonian flow.

\begin{figure*}
	\centering
	\subfloat[average face image]{\includegraphics[scale=1]{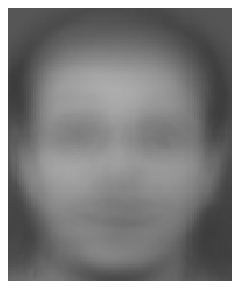}}  \hspace{ 2 cm}
	\subfloat[Face Intensity Landscape]{ \includegraphics[scale = 0.25]{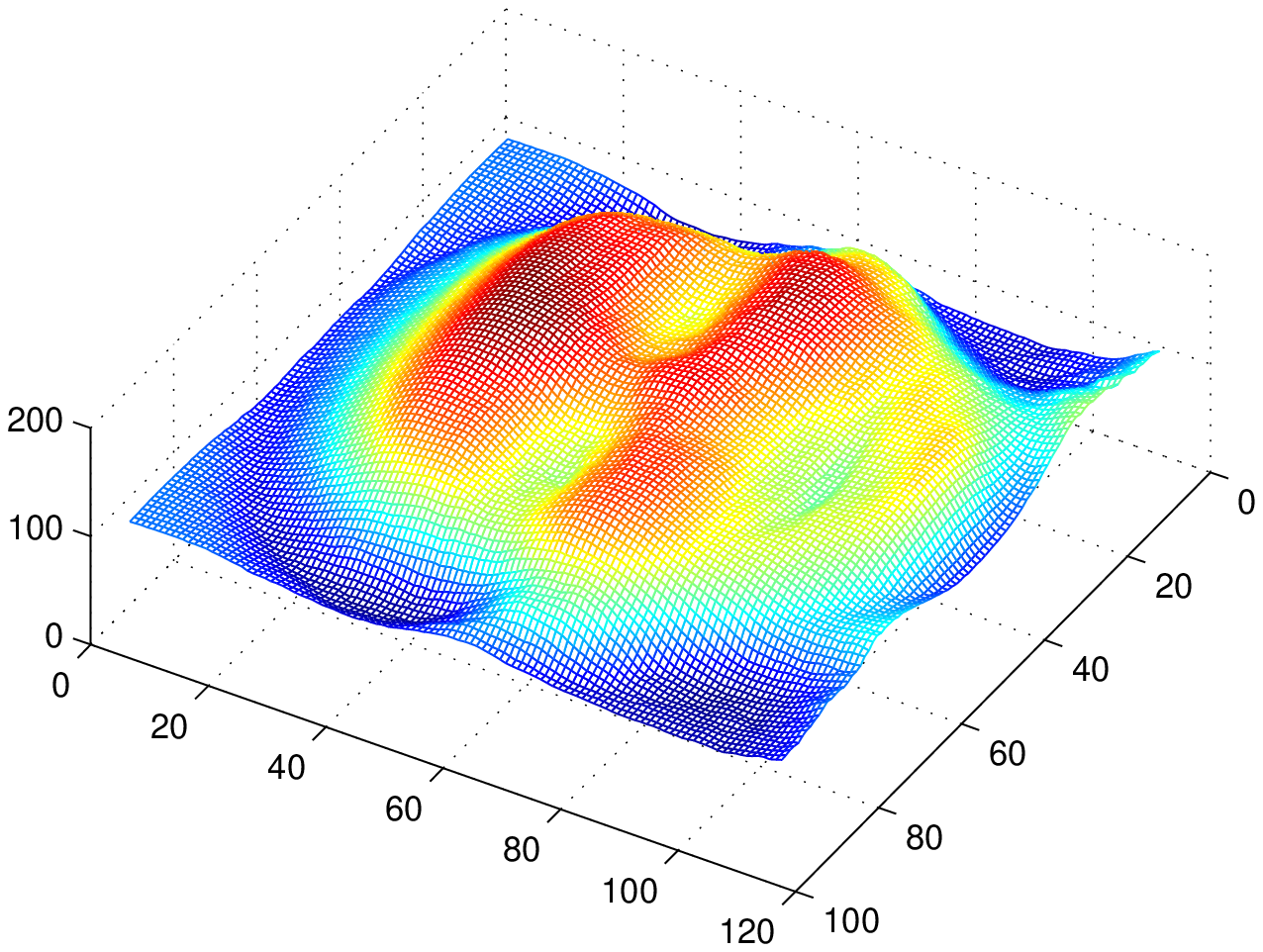} }\\
	\subfloat[Negative Gradient Flow]{ \includegraphics[scale=0.7]{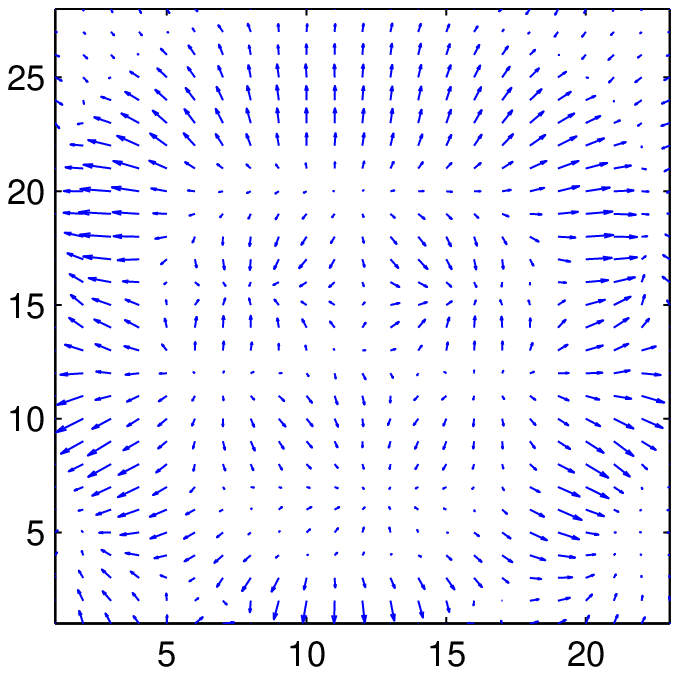} }
	\subfloat[Hamiltonian Flow]{ \includegraphics[scale=0.7]{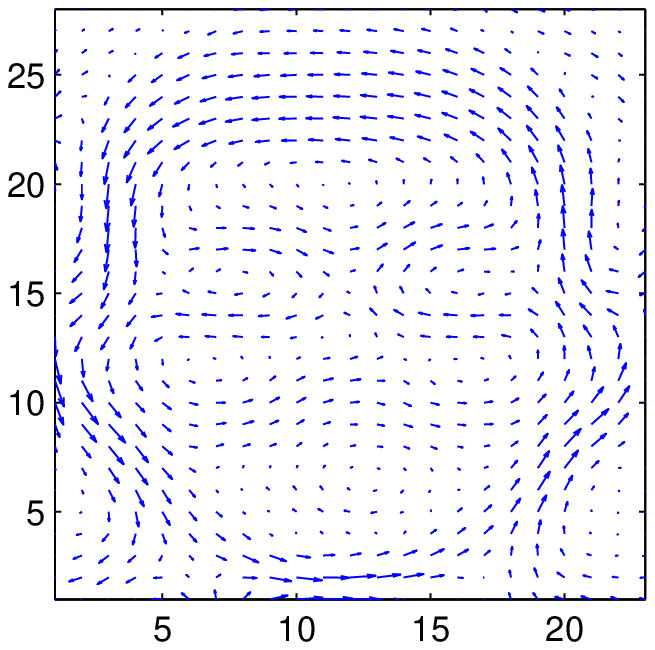} }
	\caption{Example of intensity landscape and the two dynamic systems}
	\label{fig:landscape}
\end{figure*}


\end{subsection}

\begin{subsection}{Singular Point, Attractor and Repellor}
A set is an \textbf{invariant set} if it is a union of solution curves. Singular points and periodic orbits are important invariant sets in a dynamical system. Typical singular points in a gradient system are sinks, sources, and saddles. Typical singular points in a Hamiltonian system are centers and saddles. Attractors and repellors are generalized concepts of sinks and sources. A region $R$ is attracting if for any point $x \in \partial R $, $x \cdot t \notin R $ for any $ t  < 0$. (Here, $x \cdot t$ means start with point $x$, follow the flow for time $t$).  Similarly, $R$ is repelling if for any point $x \in \partial R $, $x \cdot t \notin R $ for any $ t  > 0$ . Intuitively, for intensity landscape, one can identify valleys as attractors and ridges as repellors. In fact, ridge detection has been discussed in literature extensively. See \cite{lindeberg1998feature} for example. However, one limitation of direct ridge detection is that it is highly sensitive with respect to random perturbation. In \cite{lindeberg1998feature}, scale-space theory is used to overcome some problems.\\

In our method, we do not explicitly look for the exact locations of these invariant sets. Instead, we look for certain curves or neighborhoods that contain these invariant sets. By using tools in dynamical systems, one can predict the existence and the type of invariant sets inside the curves or neighborhoods. Poincare index and Conley index are two nature choices for this purpose. These indexes are invariant under certain deformation and perturbation of the underlying dynamical systems. We shall review basic results of these two indexes below.

\end{subsection}

\begin{subsection}{Poincare Index}

\begin{definition}
Let
\begin{equation*}
\left\{
\begin{array}{rl}
\dot{x} & = f(x,y) \\
\dot{y} & = g(x,y) 
\end{array} \right.
\end{equation*}
be a planar system. Let  $C$ be a simple closed counterclockwise curve that not passing through any singular point. Then the \textbf{Poincare Index} of $C$ is defined as 
\[ \frac{1}{2 \pi} \int _{C} d \, \{arctan( \frac{dy}{dx} ) \} \]

\end{definition}

The Poincare index has a simple geometric interpretation: consider the orientation of the vector field at a point $p \in C$. Let $p$ traverse $C$, the vector $( f(x,y), g(x,y) )$ rotates continuously and, upon returning to the original position, must have rotated through an angle $2 \pi k$ for some integer $k$, and this $k$ is the Poincare index. Cf. Figure \ref{poincareRobust}.\\

The following theorem shows how Poincare index can be used to detect and classify singular points enclosed by a curve. See \cite{andronov1971theory} for a proof.

\begin{theorem}\
\begin{itemize}
	\item[(i)] The index of a sink, a source or a center is $+1$
	\item[(ii)] The index of a hyperbolic saddle point is $-1$
	\item[(iii)] The index of a closed orbit is $+1$
	\item[(iv)] The index of a closed curve not containing any fixed points is $0$.
	\item[(v)] The index of a close curve is equal to the sum of the indexes of the fixed points within it.
\end{itemize}
\end{theorem}

\begin{remark}{Robustness of Poincare Index:}\\
The Poincare index is stable under certain deformations. In figure \ref{poincareRobust}, the vector field is deformed. However, the circle's Poincare index remains +1.
\end{remark}

\begin{figure}
	\centering
	\subfloat[]{ \includegraphics[scale=0.3]{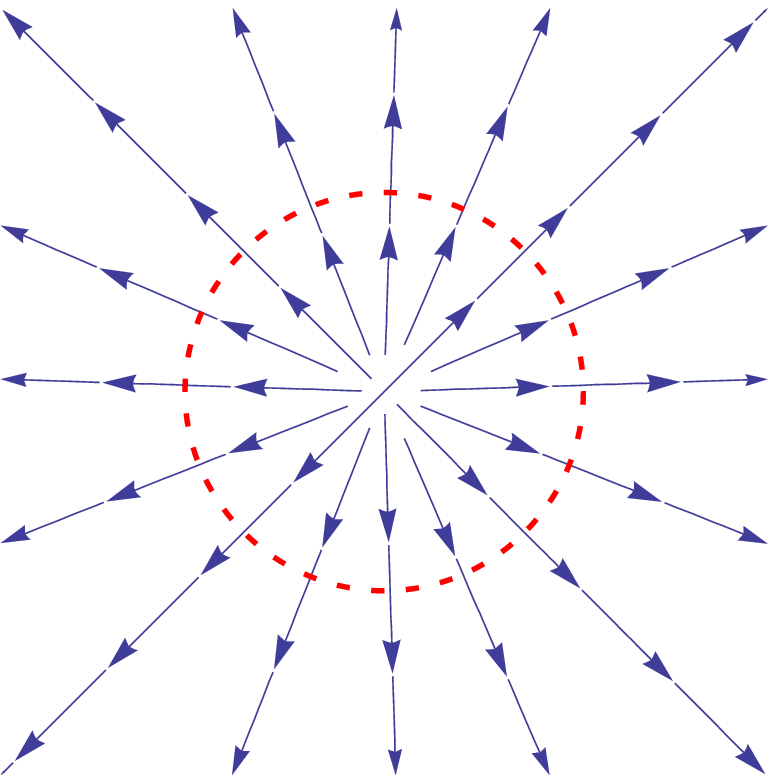} } \hspace{2 cm}
	\subfloat[]{ \includegraphics[scale=0.3]{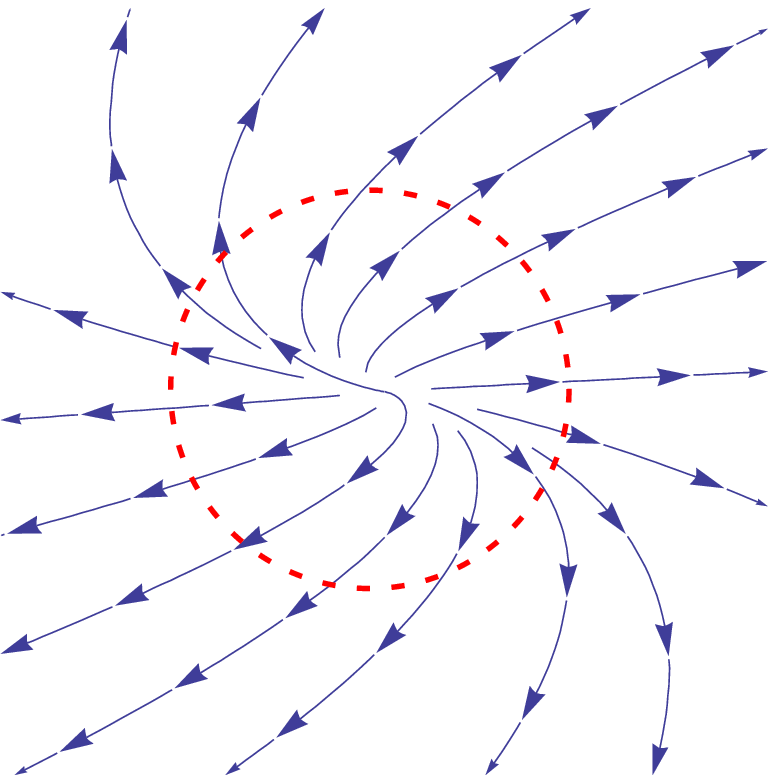} }
	\caption{Robustness of Poincare index: start with any point on the circle and traverse circle counterclockwisely. Upon returning to the starting position, the velocity vector rotates continuously for $2 \pi$. Therefore, Poincare index of the circle is +1 in both cases.}	
	\label{poincareRobust}
\end{figure}

\end{subsection}

\begin{subsection}{Conley Index}
Conley index is a powerful tool in dynamical systems. A fundamental theorem states that for a non-degenerate dynamical system, one can infer the structure of invariant sets within an so-called isolating blocks $N$ by only studying the vector field along the boundary $\partial N$. Specifically, one computes the Conley index of $N$. Moreover, the index is invariant under certain deformation of the vector field. In image pattern recognition, one important task is to identify invariance of objects from a certain class. In many cases, these invariance are also invariant sets in related dynamical systems. However, due to various factors, the intensity landscape can deform or perturbate from sample to sample. Therefore we need a robust method to detect these invariant sets. We believe Conley index is a possible solution. Next, we review basic concepts and results in Conley index theory. For more details, see \cite{mischaikow2002conley,conley1978isolated}.

\begin{definition}
A compact set $N$ is an \textbf{isolating block} iff every boundary point leaves $N$ immediately in forward time or backward time.
\end{definition}

See Figure \ref{iso_noniso} for example.

\begin{figure}
	\centering
	\subfloat[]{\includegraphics[scale=0.3]{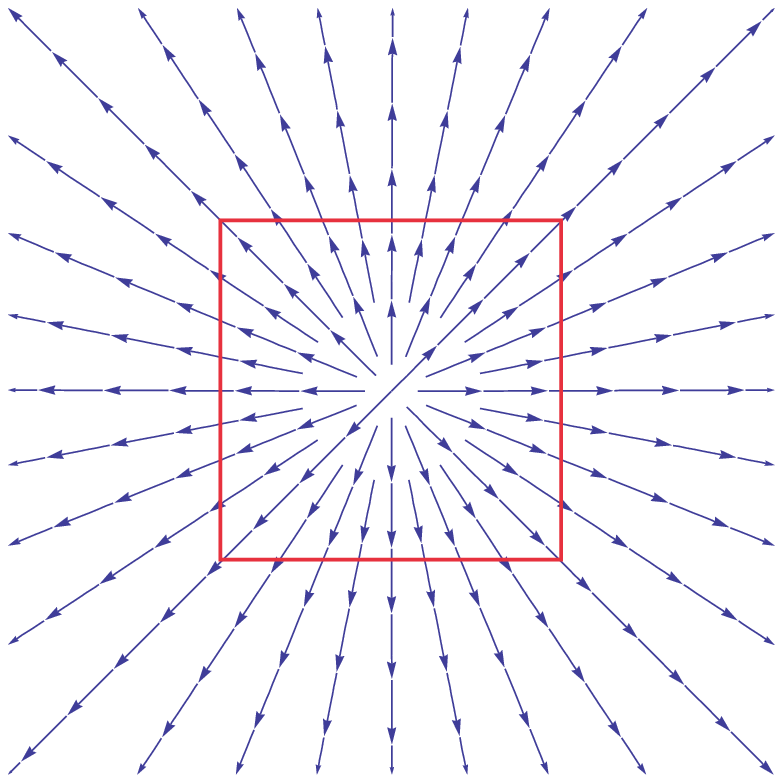}} \hspace{2 cm}
	\subfloat[]{\includegraphics[scale=0.3]{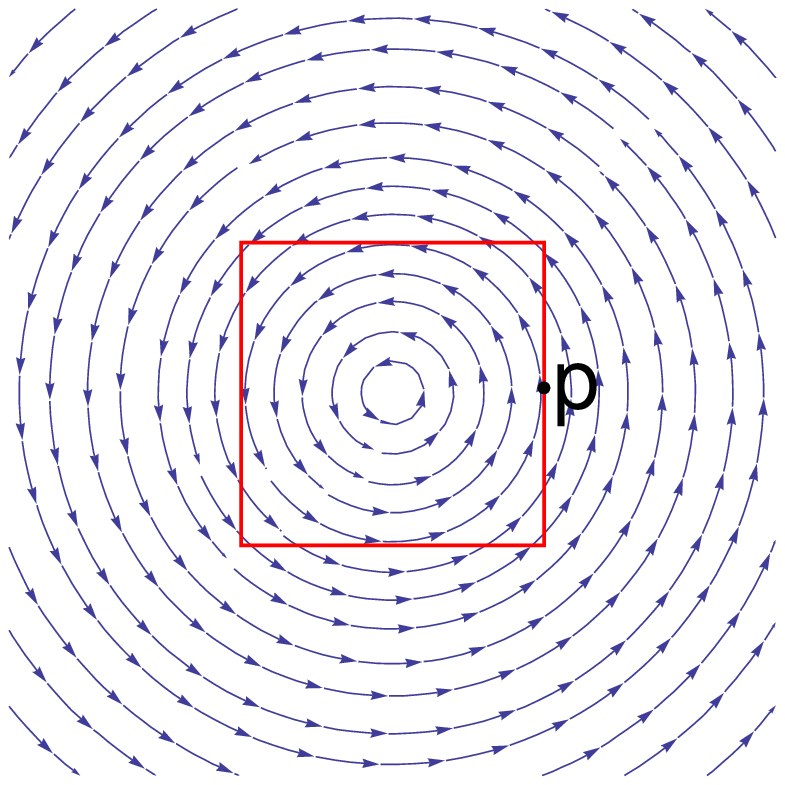}} 
	\caption{Two vector fields and their solution curves. In (a), the square is an isolating block. In (b) the square is not, because $p$ is on the boundary and the solution curve passing through $p$ is completely contained in the square.}
	\label{iso_noniso}
\end{figure}

\begin{definition}
Let $X$ be the phase space with flow $\phi$. Let $N$ be a compact set. The \textbf{exist set} $N^{-}$ is defined to be
\[ N^{-}:= \{ x\in \partial N : \phi(x,t) \notin N \quad \forall t>0 \} \]

\end{definition}

\begin{definition}
Let $N$ be an isolating block, and $N^{-}$ be the exit set. The \textbf{Conley index} is defined to be the homotopy type of the quotient space $N/N^{-}$.
\end{definition}

\begin{example}
In planar flow, the Conley index for a source is $S^2$, the 2-dimensional sphere. To see this, take $N$ to be a disk around the source, then the boundary circle $\partial N $is the exit set $N^-$. Recall that the quotient space $N/ \partial N$ is obtained by identifying $\partial N$ to a point, which gives a sphere. Cf. Figure \ref{conley_singular}. One can show that Conley index for a saddle is $S^1$, and a sink has index $S^0$. For details, see \cite{mischaikow2002conley}.

\begin{figure}
	\centering
	 \includegraphics[scale=0.4]{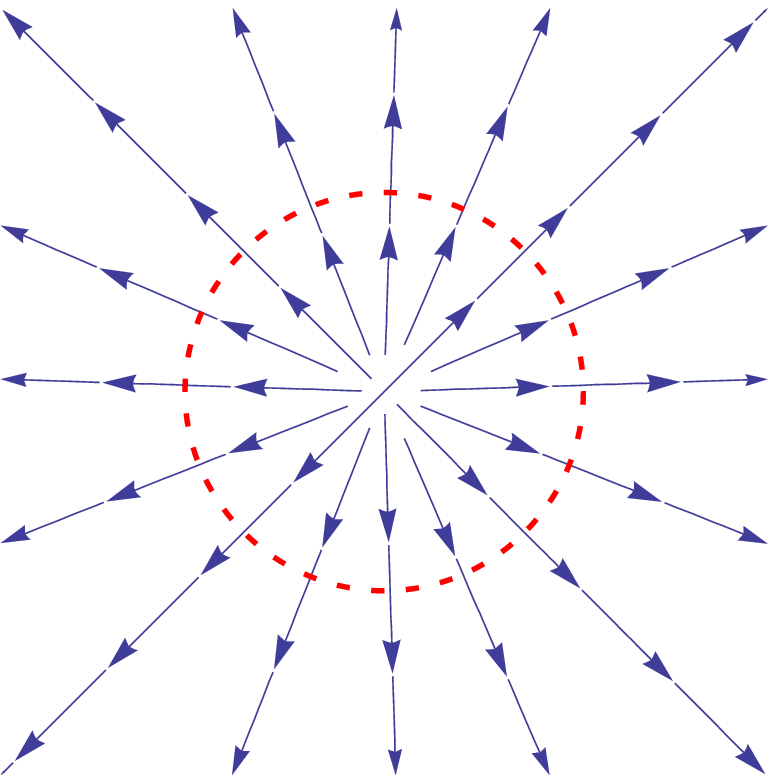}
	\caption{Conley index for a source}
	\label{conley_singular}
\end{figure}

\end{example}


\end{subsection}

\end{section}

\begin{section}{Hamiltonian Streamline Based Features}
Given an image $I(x,y)$, we can derive a negative gradient flow system and a Hamiltonian system based on $I$. We believe that solution curves of the Hamiltonian system and invariant sets in the gradient flow system are important features of the image. For example, in face detection, eyes can be identified as attracting sets and nose as repelling sets. In this section, we propose four features, which are built based on orbits in the Hamiltonian system induced by image landscape $I(x,y)$. Let $o$ be an orbit in the Hamiltonian system, we have

\begin{itemize}
	\item[1.] Density match. Let $ d(o,I) $ = $I \restriction _{o} $, i.e., the density landscape restricted on orbit $o$. If $o$ is of length $n$, then $d(o,I)$ can be identified as a point in $\mathbb{R}^n$. For any other image $\tilde{I}$, one can measure the similarity between $I$ and $\tilde{I}$ by $s(o;I, \tilde{I}) = ||d(o,I) - d(o,\tilde{I})||_2$. Suppose $I$ and $\tilde{I}$ are images from the same class, say human face, and $o$ is a feature, say the boundary of an eye. We would expect $s(o;I,\tilde{I})$ to be relatively small. In contrast, if $I$ and $\tilde{I}$ are from different classes, we expect $s(o;I,\tilde{I})$ to be large. A limitation of density match is that it may not be robust under change of lightning. Suppose $I$ and $\tilde{I}$ are human face images under different lighting conditions, then their image landscapes may have different overall latitude, this may result in a large $s(o;I,\tilde{I})$ value.
	\item[2.] Direction match. Let
		\begin{equation*}
		\left\{
		\begin{array}{rl}
		\dot{x} & = - \frac{\partial I}{\partial x} \\
		\dot{y} & = -\frac{\partial I}{\partial y} 
		\end{array} \right.
		\end{equation*}
	be a negative gradient flow system. By normalizing this vector field, one can associate the image $I$ with a direction field, $v_{I}(x,y)$. One can identify the direction field as a map $ \phi_I : \mathbb{R}^2 \rightarrow \mathbb{R}$, where $\phi(x,y) = arctan(\frac{dy}{dx})$. Let $dir(o,v_{I}) = \phi_{I} \restriction _{o}$. For an orbit of length $n$, $dir(o,v_{I}) \in \mathbb{R}^n~$. One can define a similarity function, such that for any given image $\tilde{I}$, $s(s;I,\tilde{I}) = || dir(o,v_{I}) - dir(o,v_{\tilde{I}}) || _2$. Note, the negative gradient flow along an orbit describes relative latitude of nearby landscape around $o$. Under uniform lightening change, image landscape may shift up or down, but overall shape stays similar. In this case, direction match gives a more robust result than density match.
	\end{itemize}
	Moreover, if $o$ is a closed orbit, we have additional features
	\begin{itemize}
		\item[3.] Poincare index of $o$
		\item[4.] Conley index of the compact region enclosed by $o$
	\end{itemize}
In the next section, we will describe the methods to extract orbits in the Hamiltonian system and how to compute the Poincare index and Conley index numerically.

\end{section}

\begin{section}{Numerical Treatments}
\begin{subsection}{Solution Curves}
The very first task in our method is to extract solution curves. However, there are two difficulties: (1) the dynamical system we obtained is usually a very sparse sample of an underlying ``smooth" vector field, hence long time numerical integration may cause large error; (2) we must restrict solution curves on the lattice, therefore traditional numerical methods do not directly apply.

Since solution curves are restricted on the lattice, the question is: given current position, which of the neighborhood points should be considered as the next point along the solution curve? Here, we propose a second order Runge-Kutta like algorithm. The idea is that we first integrate half step size to obtain a middle point; then determine the final position by estimating velocity field at the middle point. In particular, we estimate the velocity at the middle point as a weighted average of velocities at nearby points. Cf. Figure \ref{forwardLattice}. This middle-point scheme is reminiscent of the second-order Runge-Kutta method, which has better accuracy than Euler's method. Empirically, our method gives satisfactory results.

\begin{algorithm}
\caption{Forward Algorthim}
\begin{algorithmic}[1]
\STATE Input: vector field $\mathbf{v}$ in x-y coordinate and current position $(i,j)$ in image coordinate.
\STATE Output: image coordinate of the next point along solution curve.
\STATE mid-point $z \leftarrow [j,i] + 0.5 \times \mathbf{v}(j,i) $
\STATE Let $p_1,p_2,p_3,p_4$ be the four nearest grid points. Set $\mathbf{v}_i \leftarrow  \mathbf{v} \restriction _{p_i}$.
\STATE Assign weights $w_i \propto \frac{1}{|| p_i -z ||^2}$. $i=1,2,3,4$
\STATE Set $\mathbf{v}_z \leftarrow  \sum _i w_i \mathbf{v}_i$
\STATE Let $p^*$ be the vertex such that the angle between $p^* - z$ and $\mathbf{v}_{z}$ is minimized. Output the image coordinate of $p^*$.
\end{algorithmic}
\end{algorithm}


%
%
%
%

\setlength{\unitlength}{1mm}
\begin{figure}
\centering
\begin{picture}(40, 40)
\vspace{10 mm}
\multiput(0,0)(20,0){3}{\circle*{2}}
\multiput(0,20)(20,0){3}{\circle*{2}}
\multiput(0,40)(20,0){3}{\circle*{2}}
\put(21,40){$p^*$}
\put(30,30){\circle*{2}}
\put(31,30){$z$}
\thicklines
\put(21,21){\vector(1,1){8}}
\put(29,31){\vector(-1,2){4}}
\put(18,14){$(j,i)$}
\end{picture}
\caption{Forward algorithm}
\label{forwardLattice}
\end{figure}
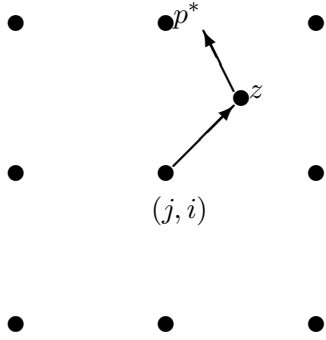

%
%
%
%

The next question is how to generate a complete orbit based on our forward algorithm. Let $\phi_t$ be a flow and $x_0$ be an initial position. Let $n$ be a non-negative integer. By $\phi_n(x_0)$ we denote the point obtained by following the flow for $n$ unit time. By $\phi_{-n}(x_0)$ we denote the point obtained by following the backward flow for $n$ unit time. A complete orbit on the lattice can be represented by 
$\{ \phi_{i}(x_0)\}_i$, where $ -m \leq i \leq n$ for some non-negative integer $m,n$.
The forward algorithm discussed previously provides an approximation for the map $\phi_1$. By applying the forward algorithm for the backward flow, we have an approximation for $\phi_{-1}$. Note that $\phi_n = (\phi_1)^n$ and $\phi_{-n} = (\phi_{-1})^n$, this gives us an iterative way to generate the complete orbit. However, we stop either we hit the boundary of the image or we detect a self-intersection. In Figure \ref{threeOrbits}, we show three orbits generated by our algorithm for the Hamiltonian system in Figure \ref{fig:landscape}. In Algorithm \ref{algo:extractAll}, we show a method to generate all solution curves for a system.

\begin{algorithm}
	\caption{Extracting all solution curves}
	\label{algo:extractAll}
	\begin{algorithmic}[1]
	\STATE Let $S$ be the set of all known solution curves. Initialize $S$ = $\emptyset$
	\STATE Start with any initial condition that is not on any existing solution curve in $S$.
	\STATE extract the complete orbit $o$
	\STATE $S$ = $S \cup o$
	\STATE Repeat 2-4 until we have exhausted all possible initial conditions. 
	\end{algorithmic}
\end{algorithm}

%

\begin{figure}
	\centering
	\includegraphics[scale=0.5]{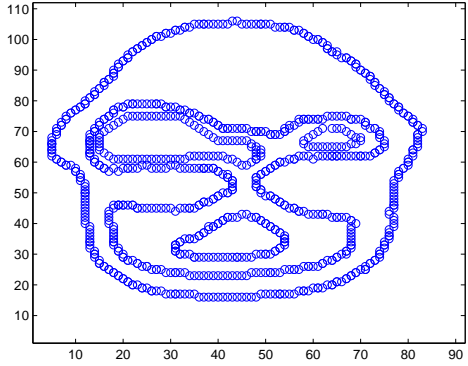} 
	\caption{Some orbits in the Hamiltonian flow generated by face image. These orbits successfully capture boundaries of eyes, nose, forehead, cheeks, mouth, etc. One should note that the dataset is small, so the intensity landscape of the average face is not entirely symmetric. This makes these orbits look imperfect.}
	\label{threeOrbits}
\end{figure}

\end{subsection}

\begin{subsection}{Numerical Treatment of Poincare Index}
Recall that the Poincare index is defined as an integration along a closed curve. Here, we are concerned with closed orbits from the Hamiltonian system and evaluate the Poincare index of these curves in the negative gradient flow system. The heuristic is that closed orbits from the Hamiltonian system usually enclose interesting regions, like sinks and sources of the associated gradient system. Due to the robustness of the index, we expect this to be a robust feature.\\

Unlike continuous flow, when working with images, we have to do integration numerically. Let $o$ be a closed orbit of length $N+1$, with $o(1) = o(N+1)$. Let $v(x,y)$ be the normalized gradient system. Then the direction field along $o$ is $\mathbf{ \phi} = v(x,y) \restriction _{o}$.  Here, each $\phi(i)$ represents the direction of the vector field, and we can identify them as angles in $[0 ,2 \pi)$. With this in mind, we compute the Poincare index as
\[ \frac{1}{2 \pi}  \sum _{i=2} ^{N+1} ( \phi(i) - \phi(i-1) ) \]
Evaluation of the above equation is straightforward. However, there is a subtlety: let $a_2$ and $a_1$ be two consecutive angles along the integration path, and we want to evaluate $a_2 - a_1$, then 

\begin{equation*}
a_2 - a_1 = \left\{
\begin{array}{rl}
a_2 + 2 \pi - a_1 & \text{if } a_2 \in [0, \pi/2], a_1 \in (3 \pi /2,2 \pi)\\
a_2 - (a1 + 2 \pi ) & \text{if } a_2 \in [3 \pi /2, 2 \pi ) , a_1 \in [0, \pi /2 ) \\
a_2 - a_1 & \text{otherwise}
\end{array} \right.
\end{equation*}
%
%

Another thing to note is we assume the closed orbit $o$ to be counterclockwise. This can be done, for example, by built-in matlab function poly2ccw(). However, this is not absolutely necessary. If $o$ is clockwise, then the index computed will be the actual Poincare index multiplied by $-1$. This will not affect the effectiveness of the feature, as long as one convention is followed.

\end{subsection}

\begin{subsection}{Numerical Treatment of Conley Index}
In general, it is highly complicated to compute the Conley index, because the index is the homotopy type of a quotient space. For those who are interested in the general method, we recommend \cite{chomp}. However, in this paper, we are only interested in planar flow, and we restrict our isolating blocks to be disks (up to homeomorphisms). One can show that under these assumptions, the Conley index is one of the following types: $D^2$ ,$S^0$, $S^2$ and $\lor _{i=1} ^ N \, S^1$ for some $N \geq 1$. Cf. Figure \ref{fig:iso_nb}. See \cite{conley1978isolated} for details.
Instead of computing the index rigorously, we propose two alternative approximations, which we call continuous pseudo Conley index and discrete pseudo Conley index.
\begin{itemize}
	\item[1.] continuous pseudo Conley index: Given an isolating block $N$ and its boundary $\Gamma = \partial N$, Let $N^{-} $ be the set of exiting pixels, i.e., pixels at which the vector field pointing outwards. We define the continuous pseudo Conley index to be $card(N^{-})/ card(\Gamma)$. Note, if the true index is $S^0$, this ratio is 0; if the true index is $S^2$, this ratio is 1. However, if the true index is $\lor _{i=1} ^ N \, S^1$, such pseudo index has no virtual connection with the true index.
	\item[2.] discrete pseudo Conley index: Let $N$ be the number of connected exiting components on the boundary. Then the discrete pseudo Conley index is
	\begin{itemize}
	\item[(i)] $D^2$, if there are exactly one connected exiting component and one connected entering component
	\item[(ii)] $S^0$, if there is no connected exiting component
	\item[(iii)] $\lor _{i=1} ^ {N-1} S^1$ if there are $N \geq 2$ connected exiting components
	\item[(iv)] $S^2$ if there is one connected exiting component and no pixel is an entering pixel
	\end{itemize}

\end{itemize}

\begin{figure}
  \centering
  \subfloat[$D^2$]{\includegraphics[width=0.15\textwidth]{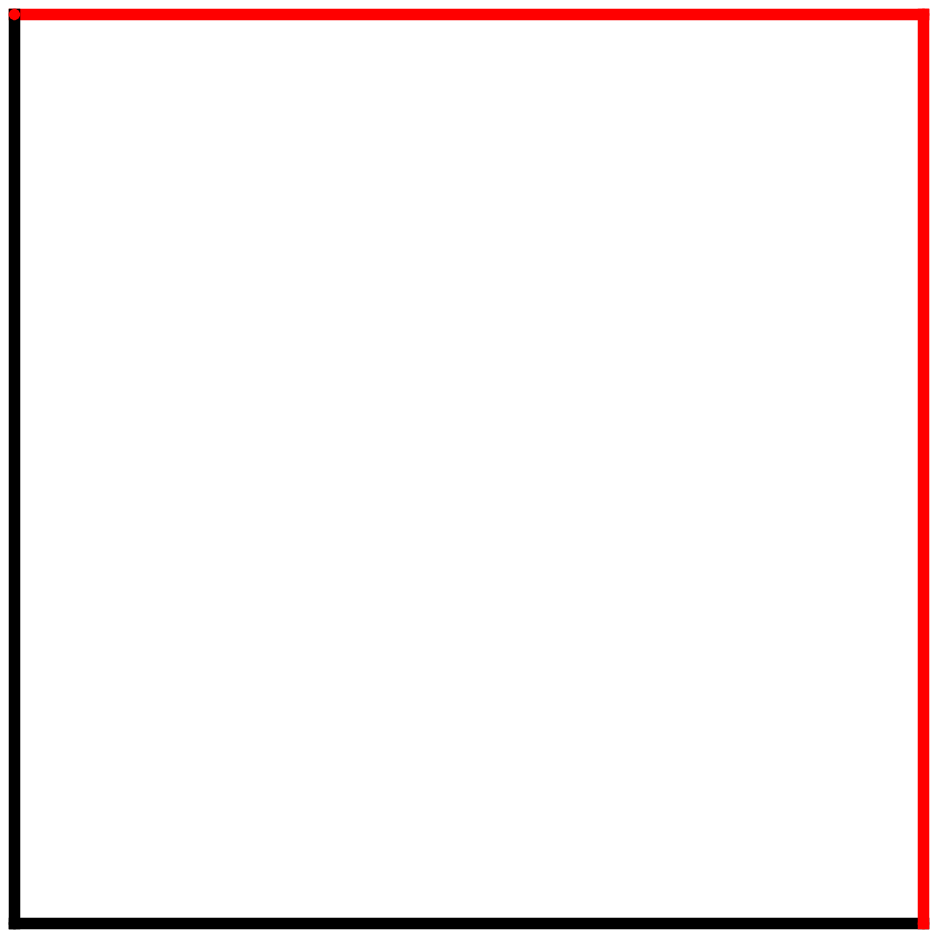}}                
  \subfloat[$S^0$]{\includegraphics[width=0.15\textwidth]{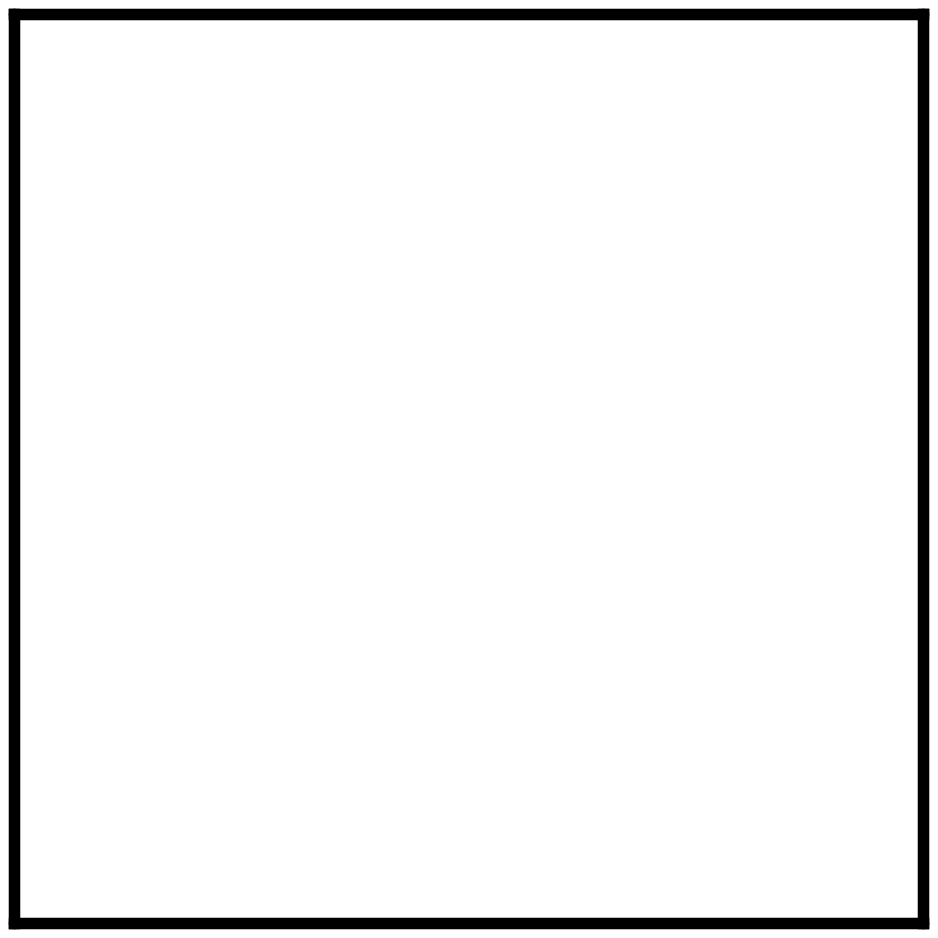}}
  \subfloat[$S^1$]{\includegraphics[width=0.15\textwidth]{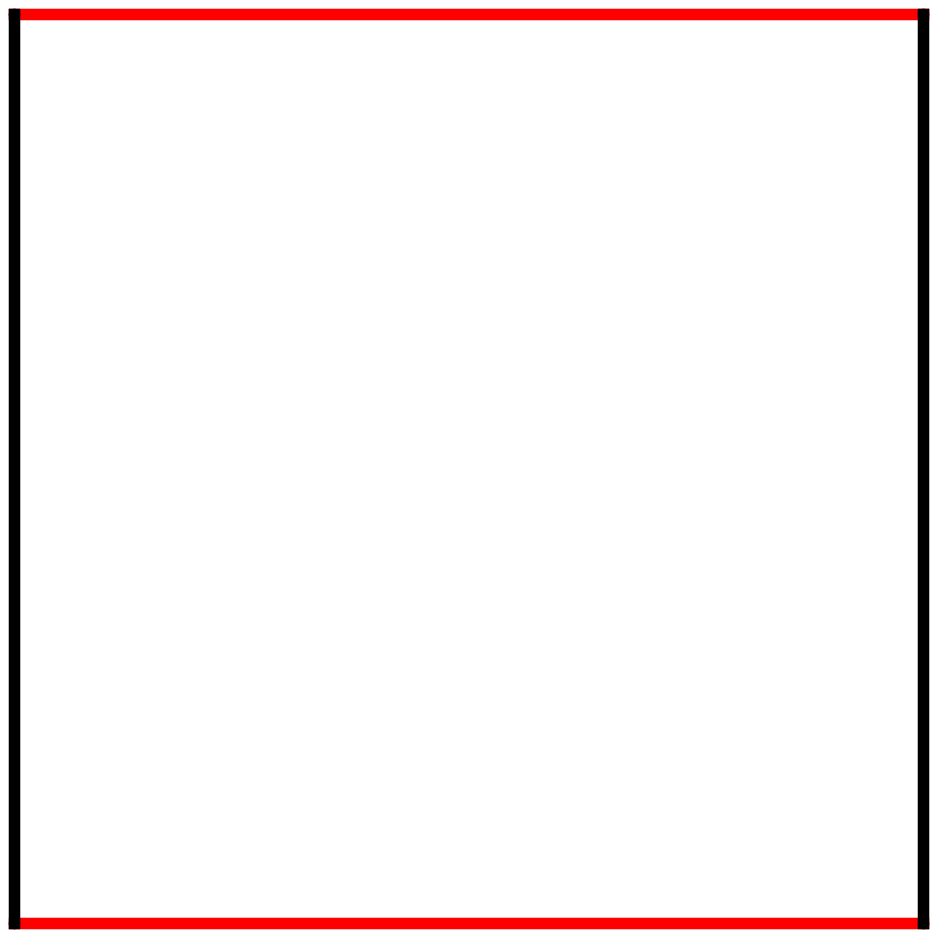}}
  \subfloat[$S^1 \vee S^1$]{\includegraphics[width=0.15\textwidth]{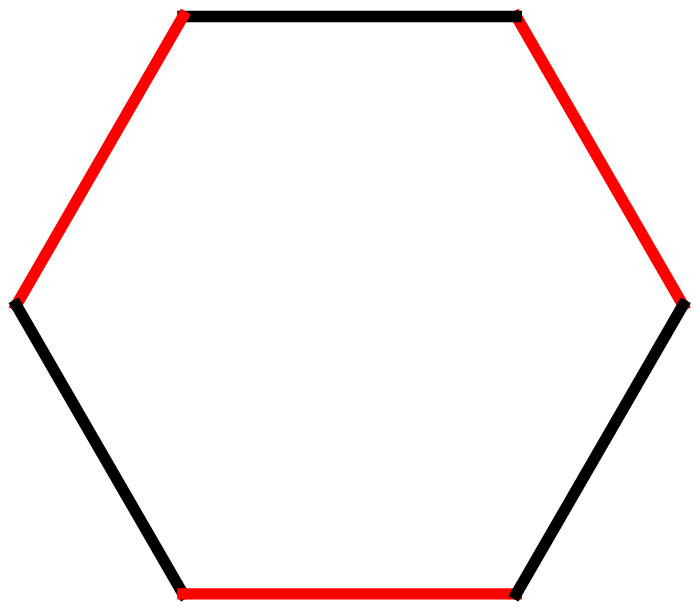}}
  \subfloat[$S^2$]{\includegraphics[width=0.15\textwidth]{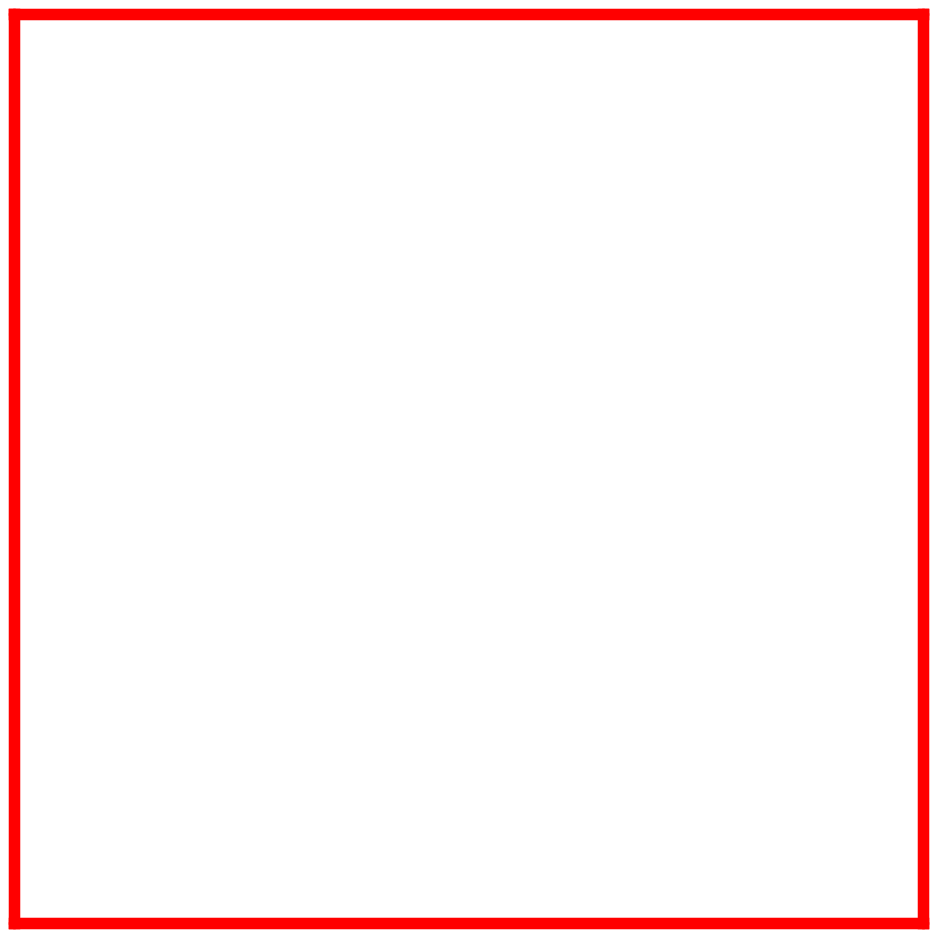}}
  \caption{Examples of isolating blocks and exiting sets for planar flow. Red edge represents a exiting set. By identifying exiting set to one point, we obtain a quotient space whose homotopy type is the Conley index.(View in color).}
  \label{fig:iso_nb}
\end{figure}

In this work, we choose to compute the continuous pseudo Conley index, because this pseudo index recovers the true index in case of attractor and repellor. We are more interested in regions of attracting and repelling, and these index serve our purpose sufficiently. 

\end{subsection}
%
%
%

\begin{section}{General Framework}

\begin{subsection}{Canonical image}
Our method relies on the hope that we can actively extract possible features of a given object class. Such features are built on two dynamical systems induced by the image landscape. However, it is not favorable to work on a single sample image, since features extracted this way may not be representative. Instead, we prefer to work on a canonical image, which provides a ``standard" image landscape for this object class. We extract feature lines from this canonical image, and use them as feature templates. We expect that images from the same class have similar feature values.

One question is how to obtain such a canonical image. In general, this can be highly nontrivial. However, if the object class has non-deformable shape and training images are well aligned, one can use the average image as a canonical image. In Figure \ref{fig:faces}, we show the average image and a random sample in our training set. The landscape of the average image is more desirable than that of the random sample.

\begin{figure}
\centering
\begin{tabular}{cc}
\includegraphics[scale = 1]{att_avg}  & \includegraphics[scale = 1]{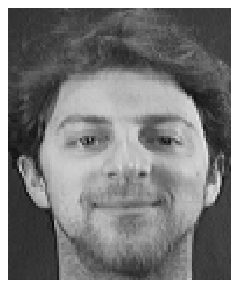} \\
\includegraphics[scale=0.25]{att_avg_landscape}& \includegraphics[scale=0.25]{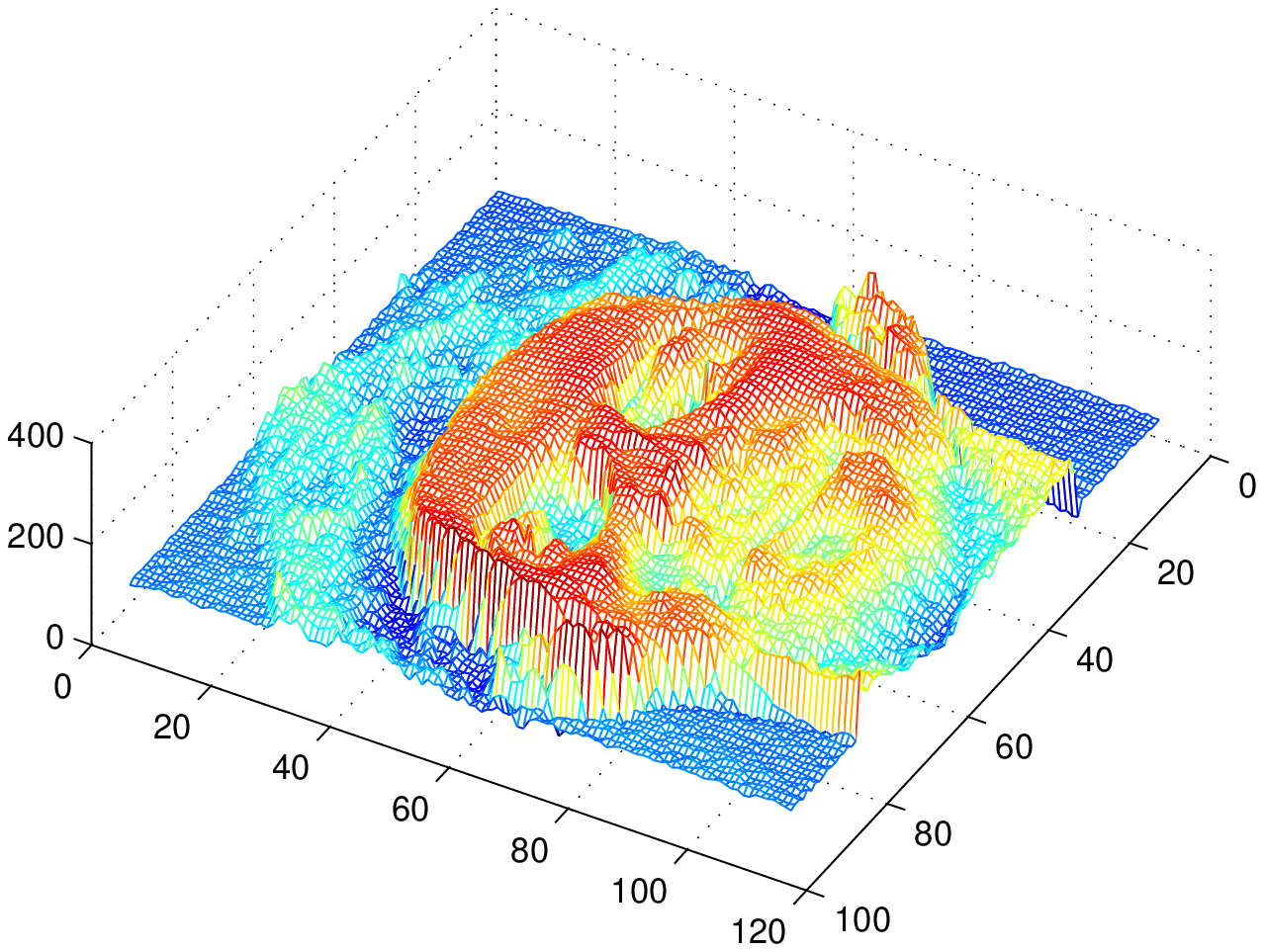} \\ 
\end{tabular}
\caption{Average face and a random sample of face. Left is the average face image}
\label{fig:faces}
\end{figure}

Our general framework has two phases: I) Design a family of features based on a canonical image. II) Use these features as weak classifiers and apply AdaBoost \cite{viola2002robust} to obtain a final strong classifier. We refer readers to \cite{viola2002robust} for the AdaBoost algorithm. Here, we give a brief outline for phase I): 
\begin{itemize}
	\item[1.] Start with a canonical image and compute its negative gradient system and Hamiltonian system. 		     \item[2.] Extract all orbits in the Hamiltonian flow. 
	\item[3.] Any orbit is associated with at least two features: density feature and gradient direction feature. 
	\item[4.] If an orbit is closed, we associate two more features with it: the Poincare index and the Conley index.
	\item[5.] We evaluate feature values according to their types.
\end{itemize}

\end{subsection}

\begin{subsection}{Experiment Results}

\begin{subsubsection}{The Datasets}
We have done experiments on the AT\&T ORL dataset and the MIT CBCL 
dataset. (Thanks to AT\&T Laboratories Cambridge and MIT CBCL). The 
ORL dataset contains 400 face images of 40 people. Images are of size 
$112 \times 92$. To get negative samples, we randomly extracted 800 
non-face images from the Caltech-256 \cite{griffinHolubPerona} clutter set. We use 200 faces and 400 non-faces for training and the rest for testing. The MIT CBCL dataset contains images of size $19 \times 19$. Training set contains 2429 faces and 4548 non-faces, the test set containing 472 faces and 23,573 non-faces.

For both datasets, we compare our Hamiltonian classifiers with Haar-like classifiers. A comparison with SVM is also included.
\end{subsubsection}

\begin{subsubsection}{Experiment on the ORL Dataset}
We take the average of training faces, and use it as a canonical image. The intensity landscape of the model image gives us 45 orbits in the associated Hamiltonian flow, of which 20 are closed. This gives us 130 Hamiltonian streamline based features. We also generate Haar-like features of those types suggested in \cite{viola2002robust}. For training images of this size, there are about ${{92}\choose{2}}{{112}\choose{2}} \approx 2.6 \times 10^7$ possible Haar-like features. However, in our experiments, we uniformly choose about 27,000 such features.

\begin{paragraph}{Using Hamiltonian Streamline Features}
If we only use the 130 Hamiltonian streamline features, training time 
is very short: approximately 2.5 minutes on our machine (Intel(R) 
Core(TM)2 Duo CPU P8600 @ 2.40GHz 2.40GHz).  Figure \ref{threeOrbits} 
shows the first few features selected by AdaBoost. 

Since we have designed four types of Hamiltonian streamline features, one natural question is in what order these types of features are selected in AdaBoost. In our test, we find ``direction match" is always the first selected feature, followed by ``density match". Conley index comes third, while Poincare index is least favorable. In a 20-stage classifier, 6 of which are ``density match", 11 of which are gradient ``direction match", 1 of which is Poincare index, 2 of which are Conley index.


%

\end{paragraph}

\begin{paragraph}{Comparing with Haar-like Features}\label{sec:comp_haar}
One would wonder how our Hamiltonian streamline features stand against Haar-like features. If we use 27,000 Haar-like features, AdaBoost takes more than 1 hour on the same machine. Testing accuracy is slightly worse than the Hamiltonian classifier. See Figure \ref{table:hami_vs_haar}.

\begin{figure}
\begin{center}
\begin{tabular}{|c|c|c|}
\hline & Hamiltonian & Haar \\ 
\hline 10-stage classifier & 1 fn, 0 fp & 3 fn, 4 fp \\ 
\hline 20-stage classifier & 0 fn, 0 fp & 2 fn, 1 fp \\ 
\hline 100-stage classifier & not needed & 1 fn, 0 fp \\ 
\hline 
\end{tabular} 
\caption{fp = false positive, fn = false negative}
\label{table:hami_vs_haar}
\end{center}
\end{figure}

\end{paragraph}

\begin{paragraph}{Joint selection: Hamiltonian streamline features and Haar-like features }
It would be interesting to see, when pooled together, which type of feature is favored by AdaBoost algorithm: Hamiltonian streamline features or Haar-like features? We put 130 Hamiltonian features with 27000 Haar-like features and let AdaBoost do feature selection. In Figure \ref{fig:weights} we show the first 20 features selected by AdaBoost and their weights in the strong classifier. It turns out that Hamiltonian streamline features are more favorable.

\begin{figure}
	\centering
	\includegraphics[scale=0.5]{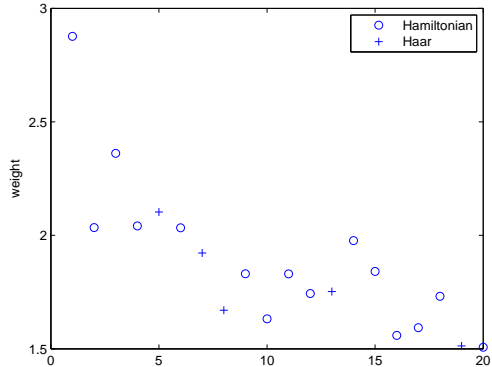}
	\caption{First 20 selected features, their types and weights in the final strong classifier}
	\label{fig:weights}
\end{figure}

\end{paragraph}

\begin{paragraph}{Detection Speed}
We have seen that with our Hamiltonian features, training speed is extremely fast. It is natural to ask whether the Hamiltonian features can achieve high speed in detection. The computational cost of evaluating Hamiltonian features is proportional (with a low constant) to the length of the orbits. In our experiment on the ORL dataset, the mean length of the orbits is 149 (median is 77). The 10-stage classifier has about the same accuracy as the 100-stage Haar-like feature classifier. In the 10-stage Hamiltonian classifer, total length of orbits is 1774. To evaluate a single Haar-like feature, we need to access 8-16 pixels in the integral image (depending on its type). In total, we need to access 800-1600 pixels. The total operations needed in two methods do not have a significant difference asymptotically. Empirically, we find both classifiers have about the same speed in classifying one subwindow. Since Haar-like classifiers can do face detection in real-time, our feature can do as well. Moreover, boosting classifiers can be trivially parallelized; with today's GPGPU, real-time detection is not an issue.

\end{paragraph}

\end{subsubsection}

\begin{subsubsection}{Experiment on the MIT CBCL Dataset}
Since the ORL dataset is relatively small, we have also done 
experiments on the MIT CBCL dataset. Just like the ORL dataset, we 
have done 3 groups of experiments: Hamiltonian features alone, 
Haar-like features alone, Hamiltonian features plus Haar-like 
features. ROC curves are shown in Figure \ref{fig:mitROC}. For this 
dataset, Haar-like classifier performs slightly better than 
Hamiltonian classifier. We believe this is due to the low resolution 
of training images whose underlying vector fields are not smooth 
enough. If we combine two types of features, the hybrid classifier 
achieves best accuracy. The ROC curve of the hybrid classifier is 
almost the same as the ROC curve of a SVM classifier using 2nd degree 
polynomial kernels \footnote{SVM result obtained from 
\url{http://vision.ai.uiuc.edu/mhyang/face-detection-survey.html}}.

\begin{figure}
	\centering
	\includegraphics[scale=0.7]{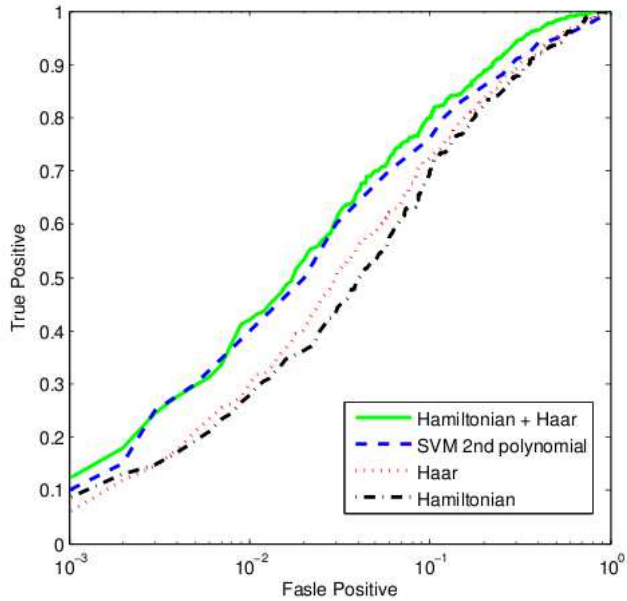}
	\caption{ROC curves on the MIT CBCL dataset}
	\label{fig:mitROC}
\end{figure}

\end{subsubsection}

\begin{subsubsection}{Discussions}

We summarize some observations in our experiments and discuss several advantages of Hamiltonian features over Haar-like features as follows:

\begin{enumerate}
	\item[(1)] Training with Hamiltonian features is much faster. Computational cost of an AdaBoost algorithm is proportional to the number of weak classifiers (features). For $n \times n$ training images, there are at least $\Theta(n^4)$ possible Haar-like features. However, empirically, our feature is of order $O(n)$. 
	\item[(2)] Hamiltonian features are more descriptive and concise. Unlike Haar-like features which are randomly generated, Hamiltonian streamline features can intelligently capture class-related features. In an AdaBoost classifier, one needs far fewer Hamiltonian weak classifiers than Haar-like feature based classifiers. This helps to keep detection speed comparable with Haar-like classifiers.
	\item[(3)] For training images with higher resolution, Hamiltonian features have better performance. For low resolution training images, Hamiltonian features may perform slightly worse, but can be used jointly with Haar-like features to achieve better accuracy than using Haar-like features alone. Since the Hamiltonian features are built based on two vector fields derived from an image, their performance depends on how well we can approximate these underlying vector fields using algorithms in Section 4. Algorithm 1 is a second order Runge-Kutta like algorithm, and the second oder Runge-Kutta method has error bound $O(1/h^2)$, where h is the step size. The ORL images have about 5 times more resolution than CBCL images (in one dimension). Thus, the step size for ORL is about 1/5 of that for CBCL. According to the error bound, orbits generated from ORL images are much more accurate. We conclude that Hamiltonian features can achieve higher detection accuracy, if higher resolution images are used.
\end{enumerate}

%

\end{subsubsection}

\end{subsection}

\end{section}

\end{section}

\begin{section}{Limitations and Future Work}
In this section, we discuss some limitations of our method and possible future works. Our method needs a canonical image. In our experiment, we take the average of training images as the canonical image. However, if training samples are not well aligned, the resulting average may be undesirable. In addition, if the object class has various possible shapes (e.g. cars), then it is not reasonable to use only one canonical image. One possible solution is to sample sufficient number of training images and use them as a family of canonical images. Then extract all features from all these images.

In addition, our method is not robust against moderate rotations, because the first two feature are not so. However, since our training time is short, one can rotate the canonical image for a series of degrees and train a series of classifiers. But this will certainly increase detection time.

\end{section}

\begin{section}{Conclusion}
This paper presents a feature extraction method based on dynamical systems induced by image landscapes. Such features capture both topological and geometrical characteristics of the image landscape of an object class, and can be used in a boosting algorithm for object detection. In our experiments, we find training with Hamiltonian features is much faster than training with Haar-like features. In addition, for training images with good resolution, Hamiltonian features also yield better accuracy. In a final strong classifier, one needs much fewer Hamiltonian weak classifiers than Haar-like weak classifiers. We find detection speed of a Hamiltonian classifier is comparable to that of a Haar-like classifier. We believe our feature can be used as a good complement for existing features in the computer vision community. 
\end{section}

%

\end{document}